\documentclass[final]{cvpr}

\usepackage{times}
\usepackage{epsfig}
\usepackage{graphicx}
\usepackage{amsmath}
\usepackage{amssymb}
\usepackage{booktabs}
\usepackage{multirow}
\usepackage{lscape}
\usepackage{graphicx}
\usepackage{color}
\usepackage[dvipsnames]{xcolor}
\usepackage[normalem]{ulem}
\useunder{\uline}{\ul}{}

\usepackage[pagebackref=true,breaklinks=true,colorlinks,bookmarks=false]{hyperref}

\def\cvprPaperID{7043} 
\def\confYear{CVPR 2021}

\begin{document}

\title{Open-book Video Captioning with Retrieve-Copy-Generate Network}

\author{Ziqi Zhang\textsuperscript{1,2,3}, 
	Zhongang Qi\textsuperscript{2},
	Chunfeng Yuan\textsuperscript{1}\thanks{Corresponding author.},
	Ying Shan\textsuperscript{2},
	Bing Li\textsuperscript{1},
	Ying Deng\textsuperscript{5},	
	Weiming Hu\textsuperscript{1,3,4}\\
	\textsuperscript{1}NLPR, Institute of Automation, Chinese Academy of Sciences\\
	\textsuperscript{2}Applied Research Center (ARC), Tencent PCG\\
	\textsuperscript{3}School of Artificial Intelligence, University of Chinese Academy of Sciences\\
	\textsuperscript{4}CAS Center for Excellence in Brain Science and Intelligence Technology\\
	\textsuperscript{5}School of Aeronautical Manufacturing Engineering, Nanchang Hangkong University\\
	{\tt\small
		\{ziqi.zhang,cfyuan,bli,wmhu\}@nlpr.ia.ac.cn,\{zhongangqi,yingsshan\}@tencent.com}
}

\maketitle

\begin{abstract} 
In this paper, we convert traditional video captioning task into a new paradigm, \ie, Open-book Video Captioning, which generates natural language under the prompts of video-content-relevant sentences, not limited to the video itself. To address the open-book video captioning problem, we propose a novel Retrieve-Copy-Generate network, where a pluggable video-to-text retriever is constructed to retrieve sentences as hints from the training corpus effectively, and a copy-mechanism generator is introduced to extract expressions from multi-retrieved sentences dynamically. The two modules can be trained end-to-end or separately, which is flexible and extensible. Our framework coordinates the conventional retrieval-based methods with orthodox encoder-decoder methods, which can not only draw on the diverse expressions in the retrieved sentences but also generate natural and accurate content of the video. Extensive experiments on several benchmark datasets show that our proposed approach surpasses the state-of-the-art performance, indicating the effectiveness and promising of the proposed paradigm in the task of video captioning.
\end{abstract}

\vspace{-0.5cm}
\section{Introduction}
Video captioning is one of the most important vision-language tasks, and it seeks to automatically describe what has happened in the video according to the visual content. Recently, many promising methods~\cite{Zhang2020,Pan2020,Shi2020,Yuan2020,Chen2020} have been proposed to address this task. These methods mainly focus on learning the spatial-temporal representations of videos to fully tap visual information and devising novel decoders to achieve visual-textual alignment or controllable decoding.
In general, there exist some drawbacks for most of the existing work: first, since the video content is the only source of input, the generation process lacks appropriate guidance, resulting in the generations of more generic sentences; second, the memory or knowledge domain of the model is fixed after training and cannot be expanded or revisited unless retraining.

\begin{figure}
	\centering
	\includegraphics[width=0.9\linewidth]{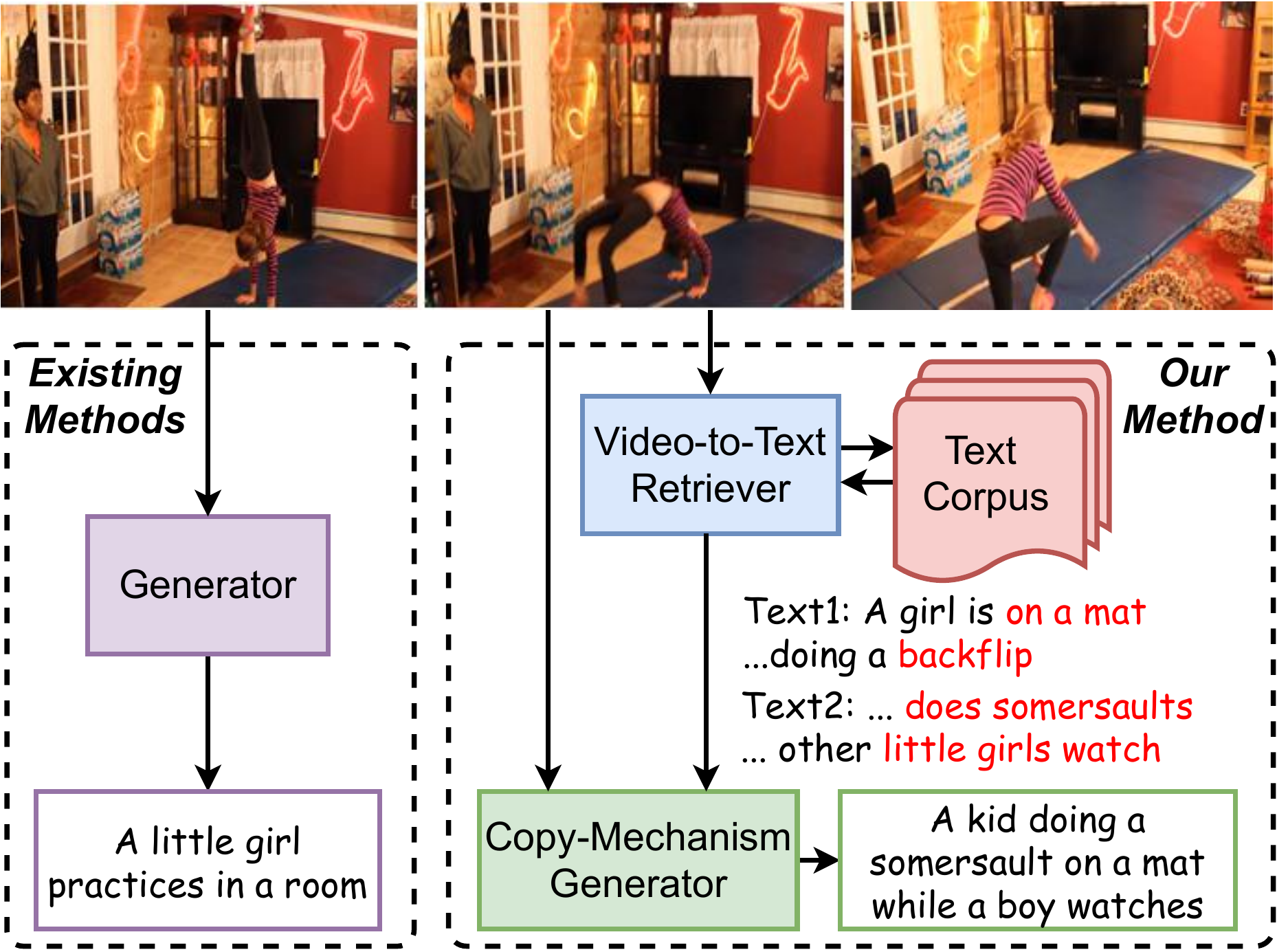}
	\label{fig:general_pipline}
	\caption{\footnotesize{Pipeline comparison of the existing methods and our method. Our generation is produced based on not only video content but also the cues of multi-retrieved sentences searched from text corpus by a cross-modal retriever. Pluggable retriever provides guidance and expansion for the generating model.}}
	\vspace{-0.5cm}
\end{figure}

To address these issues, we propose an \textit{Open-book Video Captioning} paradigm. We first compare the two cross-modal tasks for better illustration: Video-Text Retrieval (VTR) and Video Captioning (VC). VTR is a discriminative task that can access all the information of visual and textual modalities all the time; VC as a generative task can only produce words based on current generated words and visual information, which is more challenging than VTR. Instead of performing the VC task directly, we propose to convert it into two-stages: we first perform VTR to search for sentences relevant to the given video from the text corpus; then, we leverage the retrieval sentences as extra hints or guidance for caption generation. During the inference, the generator can generate words based on the video content or directly copy expressions from retrieved sentences. The flexible VTR and the changeable corpus provide the possibility for the model's extension or revision.

The inspiration for the proposed paradigm comes from the Open-domain Question Answering task~\cite{Lewis2020,Guu2020,Karpukhin2020}, which requires a system to answer any questions utilizing large-scale documents. \textit{Open} means not providing the system with documents containing the correct answers directly but requires the system to retrieve documents related to the question from massive corpus and then generate the correct answer based on them. We claim that the open-domain mechanism is also effective in cross-modal interaction area and can thus be applied to the video captioning task.

This mechanism essentially extends the knowledge domain of the model learned only from the labeled data. It is known that producing large-scale, high-quality labeled data is extremely laborious and time-consuming. Instead, a model learns to collect related references, distinguish useful hints, abstract and summarize information from external weakly labeled or unlabeled documents breaks through the limitations of labeled data. This is valuable especially for the industry-scale video platforms where hundreds of millions weakly labeled or unlabeled data are generated everyday. 
Unlike traditional semi-supervised learning that utilizes fixed weakly labeled or unlabeled examples directly for training, the proposed paradigm makes the model learn to extract useful information from changeable weakly labeled or unlabeled corpus directly for inference.

To realize the aforementioned open-book video captioning, we introduce a novel Retrieve-Copy-Generate (RCG) network. We introduce a Video-to-Text Retriever to search for video-content-relevant sentences from the corpus containing the whole sentences of the training set. Our retriever follows the Bi-encoders~\cite{Humeau2020} structure and utilizes both motion and appearance features to search desired sentences efficiently and effectively. For the example in Fig.\ref{fig:general_pipline}, the top retrieved sentences contain expressions \textit{``on a mat"}, \textit{``does somersaults"}, and \textit{``someone watches"}, which describe the given video accurately. Then, the retrieved sentences and the visual features are passed to the generator. A novel Copy-Mechanism Generator is introduced, which dynamically decides whether to copy the expressions directly from multi-retrieved sentences or generate new words from the video contents. The model combines the information from the video content and the words copied from retrievals to generate the final caption \textit{``A kid doing a somersault on a mat while a boy watches"}, which is much better than the generic caption \textit{``A little girl practices in a room"}.

The contributions of this work are listed as follows:
\begin{itemize}
	\vspace{-0.2cm}
	\item[(1)] We propose to solve the video captioning task with an open-book paradigm, which generates captions under the guidance of video-content-retrieval sentences, not limited to the video itself. 
	\vspace{-0.2cm}
	\item[(2)] We introduce a novel Retrieve-Copy-Generate network to tackle this task, where an improved cross-modal retriever is utilized to provide hints for generator, and a copy-mechanism generator is proposed for dynamical copying and better generation. 
	\vspace{-0.2cm}
	\item[(3)] The extensive experimental results highlight the benefits of combining cross-modal retrieval with copy-mechanism generation for the video caption task. The proposed approach achieves state-of-the-art results on VATEX and superior performance on MSR-VTT. 
\end{itemize} 
 
\begin{figure*}
	\small
	\centering
	\includegraphics[width=0.87\linewidth]{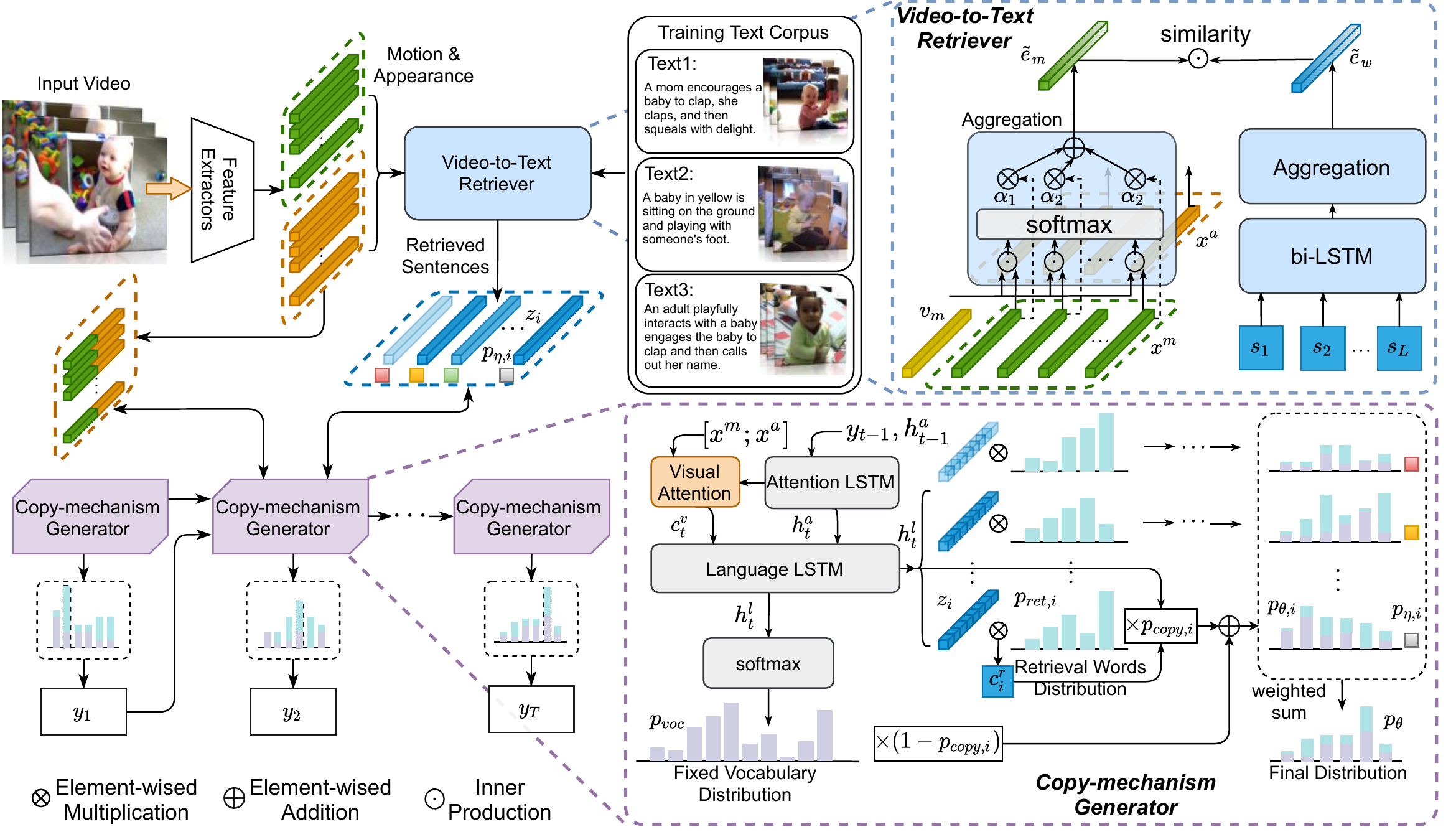}
	\caption{\footnotesize{Overview of the proposed Retrieve-Copy-Generate Network for Open-book Video Captioning. The left side is the pipeline of our method, which consists of two components: the Video-to-Text Retriever that searches for the video-content-relevant sentences from the corpus containing all the sentences in the training set; the Copy-mechanism Generator produces the words by steps under the hints or guidance of retrieved sentences and the visual features. In the upper right corner, the Bi-encoders architecture is leveraged to efficiently and effectively achieve the cross-modal retrieval. Here we only show the similarity calculation process based on the motion features. In the bottom right corner, a hierarchical caption decoder is used to generate the fixed vocabulary based on the video content. Meanwhile, an improved multi-pointer module directly copies the expressions from the retrieved sentences for a better generation.}}
	\label{fig:detailed_pipline}
	\vspace{-0.5cm}
\end{figure*} 

\vspace{-0.5cm}
\section{Related Works}
\textbf{Video Captioning.} 
Reviewing the early retrieval-based visual description methods can be roughly divided into two categories: \cite{Ordonez2011,Gupta2012,Kuznetsova2012} cast the problem as retrieval from a visual space problem and transfer the retrieved descriptions to a novel image/video; \cite{Farhadi2010,Socher2014,Hodosh2015} cast the problem as retrieval from a multimodal space problem. Although retrieval-based methods can find human-like sentences with similar semantics to the video, it is challenging to generate an entirely correct description due to limited retrieval samples. With the advent of the encoder-decoder framework, most of the current work is studying how to better use visual features~\cite{Zhang2019,Pei2019,Aafaq2019,Zhang2020,Pan2020,Shi2020} and design elaborate models~\cite{Yuan2020,Chen2020,Wang2019,Hou2019} to generate sentences directly. However, the diversity and controllability of sentences generated in this way are not satisfactory. We coordinate the classical retrieval-based method with modern encoder-decoder method to generate descriptions with diverse expressions and accurate video contents.

\textbf{Video-Text Retrieval.} 
Video-Text Retrieval is a fundamental discriminative vision-language task that helps to learn the semantic alignment of different modalities. In general, there are two kinds of model architectures, \ie, Bi-encoders~\cite{Dong2019,Gabeur2020,Liu2019,Miech2019} and Cross-encoders~\cite{Song2019,Wray2019,Yu2018,Chen2020a}. Bi-encoders map the query and candidates into a common feature space with two separate encoders. Since there is no interaction between their features, Bi-encoders are lower-accuracy but very efficient during evaluation. Compared with it, Cross-encoders yield rich interactions between query and candidates by integrating features at an early stage. This helps to gain a higher-accuracy but steep computational cost. We make efficient and effective improvements to the Bi-encoders based model, which is more applicable to our paradigm.

\textbf{Retrieval Augmented Generation Tasks.}
A series of NLP works utilize the retrieved knowledge for better generation, such as open-domain question answering (Open-QA), neural machine translation(NMT), and dialog generation. 
REALM~\cite{Guu2020} and ORQA~\cite{Lee2019} show promising results on the Open-QA task by combining masked language models with a differentiable retriever. DPR~\cite{Karpukhin2020} implements a dense embedding-based retriever to replace traditional sparse retrievers and achieves significant performance. RAG~\cite{Lewis2020} combines this learnable retriever in an end-to-end pre-trained generator. Meanwhile, SEG-NMT~\cite{Gu2018} and RER~\cite{Hossain2020} leverage retrieval to assist the NMT system. DeepCopy~\cite{Yavuz2019} also retrieves relevant unstructured sentences as external knowledge to assist dialog generation. Inspirited from these works, we extend this pattern to cross-modal generation task, \ie, video captioning, which is quite different and challenging.

\vspace{-0.2cm}
\section{Retrieve-Copy-Generate Network}
We show the overall pipeline of the proposed Retrieve-Copy-Generate (RCG) for Open-book video captioning in Fig.\ref{fig:detailed_pipline}. It consists of two components: (1) \textbf{the video-to-text retriever} $ p_\eta(z|x) $ with parameter $ \eta $, which retrieves the top-$k$ semantically similar sentences $ z $ according to video $ x $; and (2) \textbf{the copy-mechanism generator} $ p_\theta(y_t | z, x, y_{1:t-1}) $ parametrized by learnable $ \theta $, which leverages the above additional retrieved sentences $ z $, the original visual information $ x $, and the previous $ t-1 $ generated tokens $ y_{t - 1} $ to generate the current target token $ y_t $.
Formally, the conditional probability of producing caption given video for our proposed approach are defined as follows:
\begin{equation}\label{eq1}
p(y|x) = \prod_{t=1}^{T} \sum_{i=1}^{topk} \underbrace{p_\eta(z_i|x)}_{retriever} \underbrace{p_\theta(y_t|z_i,x,y_{1:t-1}}_{generator})
,
\end{equation}
where $ y $ is the target sentence with $ N $ tokens. Since a dataset usually contains videos with semantically similar content, the corresponding sentences always have similar forms or expressions. Thus, the top-$k$ retrieved sentences $z$ can provide information related to the video's content $x$ to help the generator to produce the target sentence more accurately. Meanwhile, $ p_\eta(z|x) $ can be treated as a soft-threshold, which represents the confidence of whether the generator can copy words directly from the retrieval sentences or not. 

The theoretical part is organized as follows. We introduce how to construct a simple but effective cross-modal retriever in Sec.\ref{sec:retriever}, and how to extract representations related to video content from multi-retrieved sentences to augment the generation in Sec.\ref{sec:generator}. The iterative training procedure of the whole system is described in Sec.\ref{sec:train}. 

\subsection{Effective Video-to-Text Retriever}
\label{sec:retriever}
The major function of our retriever is to find the top-$k$ most similar sentences $ z $ given video $ x $ in a massive retrieval corpus $ \mathcal{Z} $. Note that $ \left| \mathcal{Z} \right| $ can be large enough to cover all the video contents ideally (the corpus contain nearly 0.3 million pieces of sentences in our implement), and $ k $ is usually small, such as $ 1 \sim 30 $.

The video-to-text retriever applies Bi-encoders architecture: the textual encoder $ E_{txt}(\cdot) $ maps all sentences in the corpus $ \mathcal{Z} $ into $ d $-dimensional vectors and constructs a candidate dataset; the visual encoder $ E_{vis}(\cdot) $ maps the video $ x $ into a $ d $-dimensional vector as a query. The whole retrieval model is trained by metric learning, which embeds visual and textual modalities into a joint high-dimensional semantic space. Subsequently, the similarity between the video and text is defined as the dot-product of their embedding vectors:
\begin{equation}\label{eq2}
\mathrm{sim}(x, z) = E_{txt}(z) \cdot E_{vis}(x).
\end{equation}

In this way, the sentences whose embeddings are top-$k$ closest to the video embedding are sorted by the retriever. We use a simple but effective way to build these two encoders.

\textbf{Textual Encoder.}
Given a sentence $ \boldsymbol{z} = \{ s_1, \cdots, s_L \} $, each word is first input to a bi-LSTM to generate a sequence of $ d $-dimensional context-aware word embeddings $ \boldsymbol{w} = \{ \boldsymbol{w_1}, \cdots, \boldsymbol{w_L} \} $:
\begin{equation}\label{eq3}
	\begin{aligned}
		\overrightarrow{\boldsymbol{w_t}} & = \overrightarrow{\mathrm{LSTM}}(\boldsymbol{W_s} s_t, \boldsymbol{w_{t-1}};\eta_{s}),\\
		\overleftarrow{\boldsymbol{w_t}} & = \overleftarrow{\mathrm{LSTM}}(\boldsymbol{W_s} s_t, \boldsymbol{w_{t+1}};\eta_{s}),\\
		\boldsymbol{w_t} & = (\overrightarrow{\boldsymbol{w_t}}+\overleftarrow{\boldsymbol{w_t}})/2,
	\end{aligned}
\end{equation}
where $ \boldsymbol{W_s} $ is a learnable word embedding matrix and $ \eta_s $ denotes the parameters of LSTM. 

Subsequently, all the embeddings are aggregated to a single vector as the overall representation. For simplicity, we denote the aggregation function as $\mathrm{Agg}(\cdot; \boldsymbol{v})$ which utilizes the multiplicative attention mechanism, where the parameter $ \boldsymbol{v} \in \mathbb{R}^d $ can be viewed as a learnable core that gives higher weights to more discriminative features:
\vspace{-0.3cm}
\begin{equation}\label{eq4}
\mathrm{Agg}(\boldsymbol{w};\boldsymbol{v}) = \sum_{t=1}^{L}\alpha_{t} \boldsymbol{w_t},
\end{equation}
\vspace{-0.3cm}
\begin{equation}\label{eq5}
\alpha_{t} = \mathrm{softmax}(\boldsymbol{v}^T \boldsymbol{w_t}).
\end{equation}

Thus, the word embeddings are aggregated to a single vector via $ \boldsymbol{\bar{e}_w} = \mathrm{Agg}(\boldsymbol{w}; \boldsymbol{v_s}) $, where $\boldsymbol{v_s}$ is the parameters of the word aggregation function.

\textbf{Visual Encoder.}
It is assumed that appearance features $ \boldsymbol{x^a} = \{ \boldsymbol{x^a_1}, \cdots, \boldsymbol{x^a_K} \}  $ and motion features $ \boldsymbol{x^m} = \{ \boldsymbol{x^m_1}, \cdots, \boldsymbol{x^m_K} \}  $ together constitute the representation of the video $ x $. Each feature is the original feature after linear transformation to $ d $-dimensional. 
The video is usually divided into $ K $ key frames and segments. 

For visual encoder, we directly aggregate features at different moments according to their importance, since the sequential information has been involved in motion features, and appearance features only provide discrimination for the main objects in the video. We aggregate motion embeddings via $ \boldsymbol{\bar{e}_m} = \mathrm{Agg}(\boldsymbol{x^m}; \boldsymbol{v_m}) $ and appearance embeddings via $ \boldsymbol{\bar{e}_a} = \mathrm{Agg}(\boldsymbol{x^a}; \boldsymbol{v_a}) $ by reusing the aggregation function defined by Eq.\ref{eq4}, where $ \boldsymbol{v_m, v_a} \in \mathbb{R}^d $ are the parameters of two modalities' aggregation functions.

We take the average of appearance and motion similarities as the final video-text similarity:
\begin{equation}\label{eq6}
\mathrm{sim}(\boldsymbol{x}, \boldsymbol{z}) = \boldsymbol{\tilde{e}^T_w} (\boldsymbol{\tilde{e}_m} + \boldsymbol{\tilde{e}_a}  ) / 2
,
\end{equation}
where $ \boldsymbol{\tilde{e}}= \boldsymbol{\bar{e}} / ||\boldsymbol{\bar{e}}||_2 $ is the operation of $ L_2 $ normalization. 

\textbf{Pre-training and Retrieval.} The training of our cross-modal retriever follows the contrastive learning, where each positive pair $ (x^+, z^+) $ should be closer than any other negative pairs $ (x^+, z^-) $ and $ (x^-, z^+) $ in a mini-batch. The max-margin ranking loss function pushes the hardest negative pair $ \Delta $ distance away from the positive pair, as follows:
\begin{equation}\label{eq7}
\begin{aligned}
\mathcal{L}_{ret} = [\Delta + \mathrm{sim}(x^+, z^-)-\mathrm{sim}(x^+, z^+)]_+\\
+ [\Delta + \mathrm{sim}(x^-, z^+)-\mathrm{sim}(x^+, z^+)]_+
\end{aligned} 
, 
\end{equation}
where $ [x]_+ = max(x, 0)$ and $\Delta$ is the slack coefficient. The sentences belonging to other videos in the mini-batch are all negative samples of this video and vice versa. 

Due to the independent structure of Bi-encoders, the textual embeddings can be calculated offline in advance for efficient evaluation. Given video $ x $ as the query, top-$k$ most relevant retrieved sentences can be found following Eq.\ref{eq6}, and the probability of retrieved sentence $\boldsymbol{z_i}$ is estimated as:
\begin{equation}\label{eq8}
p_{\eta}(\boldsymbol{z_i}|\boldsymbol{x}) = \mathrm{softmax}(\mathrm{sim}(\boldsymbol{x}, \boldsymbol{z_i})), \ \boldsymbol{z_i} \in \{\boldsymbol{z_1},\cdots,\boldsymbol{z_{topk}}\}
.
\end{equation}

\vspace{-0.3cm}
\subsection{Copy-mechanism Caption Generator}
\label{sec:generator}
To generate captions based on the video content and the retrieved sentences, we design a novel copy-mechanism caption generator, which consists of the Hierarchical Caption Decoder and the Dynamic Multi-pointers Module. We describe the proposed generator in detail as follows: 

\textbf{Hierarchical Caption Decoder.}
It consists of attention-LSTM and language-LSTM. Formally, the attention-LSTM tries to focus on different visual features $ \boldsymbol{x} = \left[ \boldsymbol{x^m}; \boldsymbol{x^a} \right] $ according to the current hidden state $ \boldsymbol{h^a_t} $ to achieve the visual context $ \boldsymbol{c^v_t} $, where $ \left[ \cdot ; \cdot \right] $ is the concatenation of two sets of features in the feature dimension. The current hidden state of the attention-LSTM $ \boldsymbol{h^a_t} $ depends on the previous hidden state $ \boldsymbol{h^a_{t-1}} $ and generated word $ y_{t-1} $:
\begin{equation}\label{eq9}
\boldsymbol{h^a_t} = \mathrm{LSTM}(\boldsymbol{W_e} y_{t-1}, \boldsymbol{h^a_{t-1}};\theta_{a})
,
\end{equation}
\begin{equation}\label{eq10}
\boldsymbol{c^v_t} = \mathrm{Att}(\boldsymbol{h^a_t}, \boldsymbol{x}, \boldsymbol{x};\theta_v)
,
\end{equation}
where, $\mathrm{Att}(query,key,value;\theta)$ denotes standard additive attention module with parameters $\theta$ for simplify; $ \boldsymbol{W_e} $ is the word embedding matrix.

Then, the language-LSTM aggregates the current state $ \boldsymbol{h^a_t} $ and the visual context $ \boldsymbol{c^v_t} $ to generate the probability distribution of the fixed vocabulary $ \boldsymbol{p_{voc}} $ at each time step:
\begin{equation}\label{eq11}
\boldsymbol{h^l_t} = \mathrm{LSTM}([\boldsymbol{h^a_t}, \boldsymbol{c^v_t}], \boldsymbol{h^l_{t-1}};\theta_l)
,
\end{equation}
\begin{equation}\label{eq12}
\boldsymbol{p_{voc}} = \mathrm{softmax}(\boldsymbol{W_{voc}} \boldsymbol{h^l_t} + \boldsymbol{b_{voc}})
,
\end{equation}
where $ \boldsymbol{p_{voc}} \in \mathbb{R}^{1 \times V}  $; $ \theta_v, \theta_a, \theta_l $ are parameters of the visual attention module and hierarchical-LSTMs; $ \boldsymbol{W_{voc}} $ and $ \boldsymbol{b_{voc}} $ are all learnable parameters.

\textbf{Dynamic Multi-pointers Module.}
In Sec.\ref{sec:retriever}, we get the top-$k$ most similar retrieved sentences to the given video. We also leverage the bi-LSTM$(\cdot;\theta_z)$ to encode these retrieved sentences to $ \boldsymbol{z} = \{ \boldsymbol{z_1}, \cdots, \boldsymbol{z_{topk}} \} $. Each retrieved sentence $\boldsymbol{z_i}$ consists of a set of words with their embeddings $ \boldsymbol{z_i} = \{ \boldsymbol{z^1_i}, \cdots, \boldsymbol{z^L_i} \} $. 

To draw on the expressions in multiple retrieved sentences, we propose the multi-pointers module improved on Pointer-Networks~\cite{Vinyals2015} by extending the single document copying to multiple's. At each decoding step $t$, the multi-pointers module acts on each retrieved sentence $\boldsymbol{z_i}$ separately, uses the hidden state $\boldsymbol{h^l_t}$ as the query to attend $L$ words, and produces the word probability distribution $ \boldsymbol{p_{ret, i}} \in \mathbb{R}^{1 \times L} $ of the corresponding sentence:
\begin{equation}\label{eq13}
\boldsymbol{p_{ret, i}}, \boldsymbol{c^r_{i, t}} = \mathrm{Att}(\boldsymbol{h^l_t}, \boldsymbol{z_i}, \boldsymbol{z_i};\theta_r)
,
\end{equation}
where $\mathrm{Att}(\cdot)$ is the additive attention module with parameters $ \theta_r $; $ \boldsymbol{c^r_{i, t}} $ denotes the context of the retrieved sentence which is the weighted summation of $ \boldsymbol{z_i} $ by $ \boldsymbol{p_{ret,i}} $.
Since not all the words in the retrieved sentence are valid, the model needs to decide whether to copy or generate dynamically. The probabilities of copying words from each retrieved sentence $\boldsymbol{z_i}$ are determined by the semantic context of retrieved sentence $\boldsymbol{c^r_{i,t}}$ and the decoder's state $\boldsymbol{h^l_t}$: 
\begin{equation}\label{eq14}
p_{copy, i} = \sigma(\boldsymbol{W_r c^r_{i, t}} + \boldsymbol{W_lh^l_t})
,
\end{equation}

Finally, we get the generation probability distribution $ \boldsymbol{p_\theta} \in \mathbb{R}^{topk \times V} $ , which is summarized by accumulating the probability distribution of the fixed vocabulary $ \boldsymbol{p_{voc}} \in \mathbb{R}^{1 \times V} $ and the dynamic words in retrieved sentences $ \boldsymbol{p_{ret}} \in \mathbb{R}^{topk \times V} $ conditioned on copy probability $ \boldsymbol{p_{copy}} \in \mathbb{R}^{topk \times 1} $ via broadcasting. Note that $\boldsymbol{p_{ret}}$ has been extended to the same size of fixed vocabulary.
\begin{equation}\label{eq15}
\boldsymbol{p_\theta} = (1 - \boldsymbol{p_{copy}}) \boldsymbol{p_{voc}} + \boldsymbol{p_{copy}} \boldsymbol{p_{ret}}
.
\end{equation}

\subsection{Training}
\label{sec:train}
Review the definition of the open-book video captioning, the final probability of target word is jointly predicted by the similarities $ \boldsymbol{p_{\eta}} $ of the retrieved sentences and the generation probabilities $ \boldsymbol{p_{\theta}} $ with copy-mechanism by substituting Eq.\ref{eq8} and Eq.\ref{eq15} into Eq.\ref{eq1}.
Our goal is to minimize the negative log-likelihood of each target word $ y_t $:
\begin{equation}\label{eq16}
\mathcal{L}_{gen} = - \sum_{t=1}^{T} log \sum_{i=1}^{topk} \boldsymbol{p_{\eta}}(\boldsymbol{z_i}) \boldsymbol{p_{\theta,i}}(y_t)
.
\end{equation}  

These two components can be trained separately.
Assuming that given an off-the-shelf retriever, our model can directly use the retrieval results for generation. In this case, we keep the retriever fixed, only fine-tuning the generator. This provides convenience for replacing better retrievers or adapting to different datasets. 

Additionally, the retriever and generator can be jointly trained end-to-end in an iterative manner for better performance. However, updating the retriever directly during training may decrease its performance drastically as the generator has not been well trained to begin with. For a stable training, we add the ranking loss mentioned in Sec3.2 to the generation loss as a constraint \ie, $ \mathcal{L} = \mathcal{L}_{ret} + \mathcal{L}_{gen} $. Moreover, we periodically (per epoch in our work) perform the retrieval process because it is costly and frequently changing the retrieval results will confuse the generator.
\section{Experiments}
\subsection{Experimental Settings}
\textbf{Datasets.}
We carry out all the experiments on MSR-VTT and recent VATEX, which are two large-scale video-caption datasets.
The \textbf{MSR-VTT}~\cite{Xu2016} contains 10,000 open-domain video clips from YouTube website. For each clip, there are 20 human descriptions and one of 20 categories (music, sports, \etc). We follow the standard splits with 6,573 videos for training, 497 videos for validation, and 2,990 videos for testing.
The most recent \textbf{VATEX}~\cite{Wang2019a} dataset reuses the videos from Kinetics-600 that contains 41,269 video clips with 10 English text sentences. According to the official splits, the dataset is divided into 25,991 training, 3,000 validation, and 6,000 public testing.

Due to the strict quality control, VATEX has enricher, longer (average 16 tokens \textit{v.s.} 9 tokens in MSRVTT) and more accurate annotations. Therefore, we conduct and report most of the experiments on this dataset. Other experiments can be seen in the supplementary materials. 

\textbf{Evaluation Metrics.}
We use standard captioning metrics, \ie, BLEU-4, Meteor, Rouge-L, and CIDEr, to evaluate the performance of video captioning. We pay more attention to CIDEr during experiments, since only CIDEr weights the n-grams that relevant to the video content, which can better reflect the capability on producing novel expressions. Moreover, we introduce metrics in information retrieval, including Recall at K (R@K), Median Rank (MedR), and Mean Rank (MnR), to measure the performance of the video-text retrieval. R@K measures the proportion of correct targets retrieved from $ K $ samples. MedR and MnR represent the median and average rank of correct targets in the retrieved ranking list separately.

\textbf{Implementation Details.}
For the extraction of visual features, we use C3D pre-trained on Kinetics-400 and InceptionResNetV2 pre-trained on ImageNet to extract motion and appearance features, respectively. We also conduct extra experiments for VATEX with other features, \eg, the I3D motion features pre-trained on Kinetics-600 and ResNet152 appearance features pre-trained on ImageNet. All the features above are extracted from 28 key-frames/segments of video sampling at equal intervals. For the textual embedding, the sentences longer than 40 words are truncated and initialized the word embedding with GloVe by spaCy toolkit.

For the setting of retriever, the joint embedding size of Bi-encoders is 1024. We set the margin $ \Delta $ as 0.2, the batch size as 128 and the learning rate as $ 2e^{-4} $. The retriever converges in around 10 epochs, and the best model is selected from the best results on the validation. Note that during the training and testing phase of RCG, the sentences are retrieved only from the corpus of training set. Furthermore, the sentences belonging to their own video should be excluded. Otherwise, the answer will be leaked, and the training will be destroyed.

For the setting of generator, we keep the features in sync with the retriever. The hidden size of the hierarchical-LSTMs is 1024, and the state size of all the attention modules is 512. The model is optimized by Adam. The learning rate is initialized with $ 2e^{-4} $ and decayed 0.5 times every 3 epochs. The batch size is set to 64, and the training of the model can be converged with no more than 20 epochs. During the validation, the beam-search (beam size is 3) is used for a better generation. 

 
\begin{table}[]
	\caption{Performance of video-text retrieval and video captioning tasks using fixed retrievers trained by different features.}
	\label{tab:tab1}
	\resizebox{0.48\textwidth}{!}{
		\begin{tabular}{@{}clcccccccc@{}}
			\toprule
			\multirow{2}{*}{\#} & \multirow{2}{*}{\begin{tabular}[c]{@{}l@{}}\textbf{Methods}\end{tabular}} & \multicolumn{4}{c}{\textbf{Video-Text Retrieval}} & \multicolumn{4}{c}{\textbf{Video Captioning}} \\ \cmidrule(l){3-6} \cmidrule(l){7-10} 
			&  & R@1 & R@5 & MedR $\downarrow$ & MnR $\downarrow$ & C & B-4 & R & M \\ \midrule
			1 & w/o Retriever & - & - & - & - & 49.2 & 31.3 & 48.5 & 21.9 \\
			2 & Random & 0.1 & 0.1 & 7002 & 10328 & 49.1 & 31.0 & 48.6 & 21.7 \\
			3 & ResNet & 17.2 & 37.7 & 11 & 126.8 & 51.5 & 32.2 & 49.4 & 23.1 \\
			4 & I3D & 24.9 & 51.0 & 5 & 48.8 & 54.7 & 33.2 & 49.7 & 23.3 \\
			5 & ResNet+I3D & \textbf{29.4} & \textbf{56.9} & \textbf{4} & \textbf{36.5} & \textbf{56.8} & \textbf{33.4} & \textbf{50.1} & \textbf{23.6} \\ \bottomrule
		\end{tabular}
	}
	\vspace{-0.3cm}
\end{table}

\begin{table}[]
	\centering
	\caption{Performance of training the model with different numbers of retrieved sentences. The model is tested via top-10 sentences. \textit{Fixed} denotes whether the retriever is fixed or jointly trained.}
	\label{tab:tab2}
	\resizebox{0.45\textwidth}{!}{
		\begin{tabular}{@{}ccccccc@{}}
			\toprule
			\textbf{\#} & \textbf{\begin{tabular}[c]{@{}c@{}}\# Retrievals\\ Training\end{tabular}} & \textbf{Fixed} & CIDEr & BLEU-4 & Rouge-L & Meteor \\ \midrule
			1 & \multirow{2}{*}{1} & \checkmark & 55.3 & 33.4 & 50.0 & 23.5 \\
			2 &  & $\times$ & 56.3 & 33.4 & 50.0 & 23.6 \\ \cmidrule(l){2-7} 
			3 & \multirow{2}{*}{3} & \checkmark & 56.8 & 33.4 & 50.1 & 23.6 \\
			4 &  & $\times$ & \textbf{57.5} & \textbf{33.9} & \textbf{50.2} & \textbf{23.7} \\ \cmidrule(l){2-7} 
			5 & \multirow{2}{*}{5} & \checkmark & 56.3 & 33.8 & 50.1 & 23.6 \\
			6 &  & $\times$ & 56.8 & \textbf{33.9} & \textbf{50.2} & 23.6 \\ \cmidrule(l){2-7} 
			7 & \multirow{2}{*}{10} & \checkmark & 56.4 & 33.6 & 50.0 & 23.5 \\
			8 &  & $\times$ & 57.1 & 33.7 & 50.1 & 23.6 \\ \bottomrule
		\end{tabular}
	}
	\vspace{-0.5cm}
\end{table}

\subsection{Quantitative Analysis}
The core of proposed open-book video captioning is to assist the caption generation by introducing the cross-modal retrieval. We elaborate the following \textit{Q}\&\textit{A}s to better illustrate the impact open-book captioning and prove the effectiveness of our model. Unless otherwise specified, all the experiments are carried out on VATEX.

\textbf{Does the performance of the retriever affect the results?}
We leverage different features to train the retrievers to simulate retrievers with different capabilities. We report the performance of the video-text retrieval and corresponding generation using different retrievers in Tab.\ref{tab:tab1}. In lines 3 to 5. we see that the retriever with better performance can significantly improve the generation. An intuitive explanation is that a good retriever can find sentences closer to the video content and provide better expressions. Moreover, we find the results are similar between the model without retriever in line 1 and the model with a randomly initialized retriever as the worst retriever in line 2. In the worst case, the generator will not rely on the retrieved sentences reflecting the robustness of our model. 

\textbf{Does the number of retrieved sentences affect the results?}
We analyze the effect of using different numbers of retrieved sentences in training and testing phases. 
In training phase, we explore $ 1\sim10 $ sentences for training, and 10 sentences are used for testing. As illustrated in Tab.\ref{tab:tab2}, we find that a moderate number of retrieved sentences (3 for VATEX) are helpful for generation during training. This is because the retrieved sentence does not exactly correspond to the video, and the noise is also introduced with useful information. 
In testing phase, we select a well-trained model with fixed retriever trained on 3 sentences and test with various $ 1\sim30 $ sentences retrieved from training set as hints. Compared with the results in Tab.\ref{tab:tab3} lines 4 to 6, it demonstrates that more sentences may bring more hints for generation until saturation. In summary, too many retrieved sentences with noise are not conducive to model training, and a trained model can adaptively select useful cues from multiple sentences.

\begin{figure*}
	\centering
	\includegraphics[width=1\linewidth]{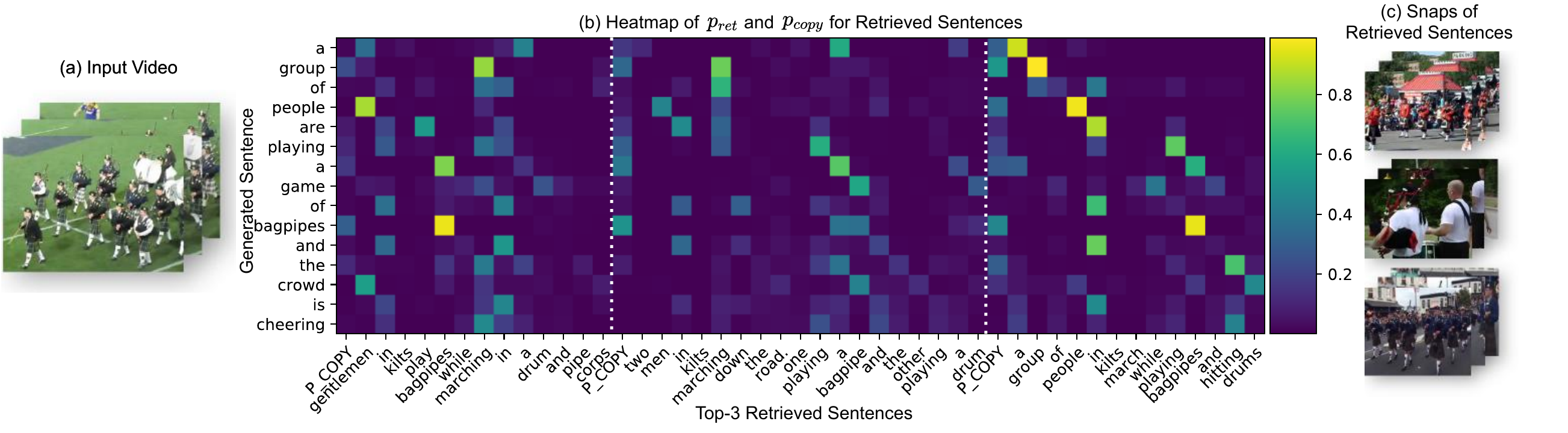}
	\caption{\footnotesize{Illustration of the words copied from the retrieved sentences and their probabilities during the generation process. Each block represents the attention weights of the words in one retrieval sentence, and the first column of each block denotes the probability of copying.}}
	\label{fig:fig3}
	\vspace{-0.5cm}
\end{figure*}

\begin{table}[]
	\centering
	\caption{Performance of testing the model with different numbers of retrieved sentences. The model is trained via top-3 retrieved sentences.}
	\label{tab:tab3}
	\resizebox{0.4\textwidth}{!}{
		\begin{tabular}{@{}cccccc@{}}
			\toprule
			\textbf{\#} & \textbf{\begin{tabular}[c]{@{}c@{}}\# Retrievals\\ Testing\end{tabular}} & CIDEr & BLEU-4 & Rouge-L & Meteor \\ \midrule
			1 & 1 & 50.5 & 30.9 & 48.5 & 22.5 \\
			2 & 3 & 54.1 & 32.6 & 49.6 & 23.3 \\
			3 & 5 & 55.8 & 33.0 & 50.0 & 23.5 \\
			4 & 10 & 56.8 & 33.4 & 50.1 & 23.6 \\
			5 & 15 & 57.4 & \textbf{33.5} & \textbf{50.2} & \textbf{23.7} \\
			6 & 30 & \textbf{57.5} & \textbf{33.5} & 50.1 & \textbf{23.7} \\ \bottomrule
		\end{tabular}
	}
	\vspace{-0.3cm}
\end{table}
         
\begin{table}[]
	\centering
	\caption{Performance of using different qualities of the corpus for testing. Different qualities are simulated through the corpus size.}
	\label{tab:tab4}
	\resizebox{0.48\textwidth}{!}{
		\begin{tabular}{@{}lllcccc@{}}
			\toprule
			\textbf{\#} & \multicolumn{1}{c}{\textbf{\begin{tabular}[c]{@{}c@{}}Retrieval\\ Corpus\end{tabular}}} & \multicolumn{1}{c}{\textbf{\begin{tabular}[c]{@{}c@{}}Corpus\\ Size\end{tabular}}} & CIDEr & BLEU-4 & Rouge-L & Meteor \\ \midrule
			1 & TrainSet & +0.1\% & 39.9 & 29.4 & 47.7 & 21.9 \\
			2 & TrainSet & +1\% & 48.9 & 31.4 & 49.0 & 22.9 \\
			3 & TrainSet & +10\% & 56.0 & 33.2 & 49.9 & 23.5 \\
			4 & TrainSet & +100\% & 56.8 & 33.4 & 50.1 & 23.6 \\
			5 & TestSet & +Oracle & \textbf{58.9} & \textbf{34.3} & \textbf{50.5} & \textbf{23.9} \\ \bottomrule
		\end{tabular}
	}
	\vspace{-0.5cm}
\end{table}
           
\textbf{Does the quality of the retrieval corpus affect the results? } 
We conduct this experiment by randomly selecting different proportions of sentences in training set to simulate retrieval corpora of different quality. In Tab.\ref{tab:tab4} lines 1 to 4, it illustrates large scale retrieval corpus is conducive to producing better generation, which may be because the higher quality corpus contains more semantically similar sentences, and it provides more hints related to the video content for generation. Furthermore, assuming that our retrieval corpus is good enough to contain sentences that correctly describe the video. It can be seen in Tab.\ref{tab:tab4} line 5, a significant improvement than ever before if we combine training set and test set as the Oracle corpus for testing. Note that the Oracle corpus is only intended to show that our model can retrieve better sentences for generation and is not involved in the training process.

\begin{figure}
	\centering
	\includegraphics[width=0.85\linewidth]{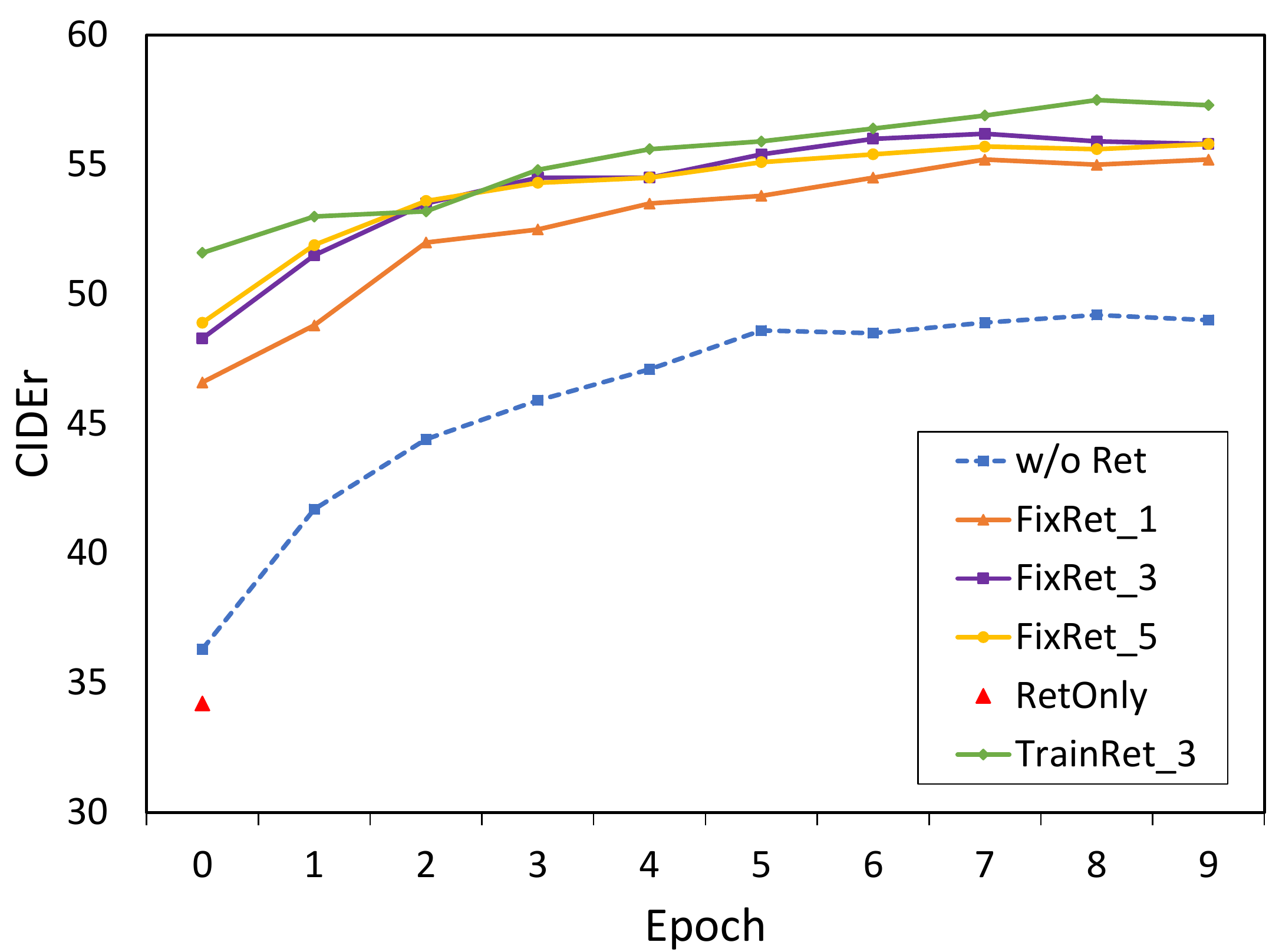}
	\caption{The CIDEr scores of different models during training on VATEX. \textit{FixRet} and \textit{TrainRet} denote the fixed retriever and jointly trained retriever separately. The number after \textit{Ret} means the number of sentences used for training. \textit{RetOnly} means directly selecting the top-$1$ retrieved sentence as the caption. \textit{w/o Ret} is our baseline model, which follows the standard generation model without our proposed retriever.}
	\label{fig:fig4}
	\vspace{-0.5cm}
\end{figure}

\textbf{Which is better, fixed or jointly trained retriever?} 
We choose CIDEr as the metric of caption performance since it reflects the generation associated with video content. We plot the real-time test scores of fixed retriever with different sentences, jointly trained retriever and \textit{without} retriever during training. As shown in Fig.\ref{fig:fig4}, the accuracy is significantly improved, and the model converges faster after introducing our retriever. The jointly trained retriever is also better than fixed in the term of accuracy. The same conclusion can be drawn from Tab.\ref{tab:tab2}. This may be because joint trained retriever can increase the probability of sentences that are more helpful for generation.

For efficiency, we profile the model speed on a server with CPU E5-2650 v4 @ 2.20GHz, 128 GB memory and GTX1080Ti. All tests are performed using a single GPU. The retrieval speed is about 254 videos/sec, which is the same for both retrievers. The fixed retriever's training speed is 137.6 videos/sec, which is slightly faster than the co-trained retriever of 127.2 videos/sec by 8.2\%. 

\begin{figure*}
	\centering
	\includegraphics[width=1\linewidth]{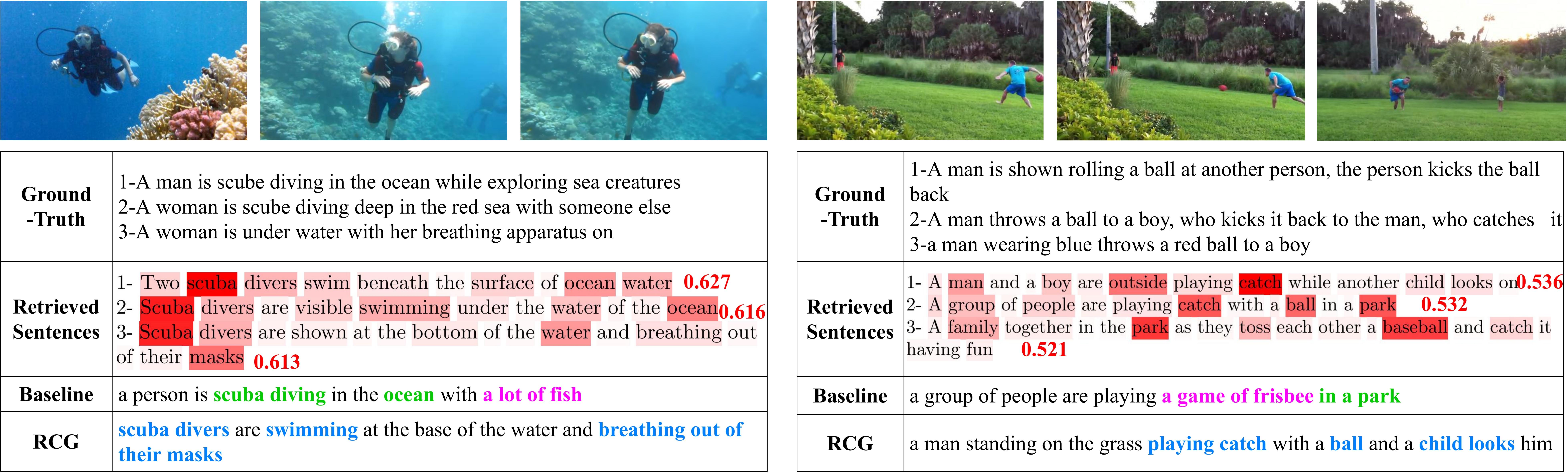}
	\caption{\footnotesize{Visualizations of the proposed RCG. The \textbf{attention weights} and \textcolor{Red}{\textbf{similarities}} between the video of top-3 retrieved sentences are shown. The baseline model can generate some \textcolor{Green}{\textbf{correct words}} and \textcolor{magenta}{\textbf{wrong words}}. Compared with it, our RCG can generate more diverse captions by copying the \textcolor{RoyalBlue}{\textbf{expressions}} from retrieved sentences.}}
	\label{fig:fig5}
	\vspace{-0.3cm}
\end{figure*} 

\begin{table}[]
	\centering
	\caption{Experiments on cross-dataset video captioning. The model is trained on VATEX training set and tested on MSRVTT test set. The sentences are retrieved from training corpus. \textit{Ret} and \textit{Gen} denote using retriever and using generator. $*$ means the retriever is trained on MSR-VTT.}
	\label{tab:tab5}
	\resizebox{0.47\textwidth}{!}{
		\begin{tabular}{llcccccc}
			\hline
			\# & \textbf{\begin{tabular}[c]{@{}l@{}}Retrieval\\ Corpus\end{tabular}} & \textbf{Ret} & \textbf{Gen} & C & B-4 & R & M \\ \hline
			1 & - & $\times$ & \checkmark & 0.159 & 0.196 & 0.463 & 0.233 \\
			2 & VATEX & \checkmark & $\times$ & 0.099 & 0.082 & 0.359 & 0.185 \\
			3 & VATEX & \checkmark & \checkmark & 0.202 & 0.217 & 0.482 & 0.242 \\
			4 & VATEX+MSRVTT & \checkmark & $\times$ & 0.123 & 0.089 & 0.363 & 0.186 \\
			5 & VATEX+MSRVTT & \checkmark & \checkmark & 0.209 & 0.222 & 0.485 & 0.245 \\
			6 & VATEX+MSRVTT & \checkmark* & \checkmark & \multicolumn{1}{l}{\textbf{0.241}} & \multicolumn{1}{l}{\textbf{0.232}} & \multicolumn{1}{l}{\textbf{0.495}} & \multicolumn{1}{l}{\textbf{0.252}} \\ \hline
		\end{tabular}
	}
	\vspace{-0.5cm}
\end{table}

\textbf{How is the generalization of the model for cross-dataset videos?} 
In practical applications, the input video distribution is not necessarily the same as that of training data. This puts forward requirements for the generalization of the model. For this experiment, we pre-train models on VATEX and measure the performances on MSRVTT that is unseen in training. As shown in Tab.\ref{tab:tab5}, the performance of our RCG in line 3 is better than the baseline generation model in line 1. The comparison to line 3,5 shows that higher quality of the retrieval corpus leads to better performance. Furthermore, we select a retriever trained on MSR-VTT, and the comparison to line 5,6 shows a better retriever can further improve performance. The above experiments also show that our RCG can be extended by changing different retriever and retrieval corpus. 

\begin{table}[]
	\caption{Performance comparison with state-of-the-art methods on MSR-VTT and VATEX testing set. +RL denotes reinforcement learning, +Audio means introducing audio features.}
	\label{tab:tab6}
	\resizebox{0.48\textwidth}{!}{
		\begin{tabular}{@{}lllcccc@{}}
			\toprule
			\multirow{2}{*}{\textbf{Dataset}} & \multirow{2}{*}{\textbf{Method}} & \multirow{2}{*}{\textbf{Ref.}} & \multirow{2}{*}{CIDEr} & \multirow{2}{*}{BLEU-4} & \multirow{2}{*}{Rouge-L} & \multirow{2}{*}{Meteor} \\
			&  &  &  &  &  &  \\ \midrule
			\multirow{11}{*}{\begin{tabular}[c]{@{}l@{}}MSR-\\ VTT\end{tabular}} & POS-CG\cite{Wang2019} & ICCV19 & 48.7 & 42.0 & 61.6 & 28.2 \\
			& POS-VCT\cite{Hou2019} & ICCV19 & 49.1 & 42.3 & 62.8 & 29.7 \\
			& SAAT\cite{Zheng2020} & CVPR20 & 49.1 & 40.5 & 60.9 & 28.2 \\
			& \quad +RL &  & 51.0 & 39.9 & 61.2 & 27.7 \\
			& STG-KD\cite{Pan2020} & CVPR20 & 47.1 & 40.5 & 60.9 & 28.3 \\
			& PMI-CAP\cite{Chen2020a} & ECCV20 & 49.4 & 42.1 & - & 28.7 \\
			& \quad +Audio &  & 50.6 & 43.9 & - & 29.5 \\
			& ORG-TRL\cite{Zhang2020} & CVPR20 & 50.9 & 43.6 & 62.1 & 28.8 \\ \cmidrule(l){2-7} 
			& Baseline & Ours & 49.8 & 42.2 & 61.2 & 28.2 \\
			& +FixRet & Ours & 52.3 & \textbf{43.1} & \textbf{61.9} & 29.0 \\
			& +TrainRet & Ours & \textbf{52.9} & 42.8 & 61.7 & \textbf{29.3} \\ \midrule \midrule
			\multirow{6}{*}{VATEX} & VATEX\cite{Wang2019a} & ICCV19 & 45.6 & 28.7 & 47.2 & 21.9 \\
			& ORG-TRL\cite{Zhang2020} & CVPR20 & 49.7 & 32.1 & 48.9 & 22.2 \\
			& NSA\cite{Guo2020} & CVPR20 & 57.1 & 31.0 & 49.0 & 22.7 \\ \cmidrule(l){2-7} 
			& Baseline & Ours & 49.2 & 31.3 & 48.5 & 21.9 \\
			& +FixRet & Ours & 56.8 & 33.4 & 50.1 & 23.6 \\
			& +TrainRet & Ours & \textbf{57.5} & \textbf{33.9} & \textbf{50.2} & \textbf{23.7} \\ \bottomrule
		\end{tabular}
	}
	\vspace{-0.5cm}
\end{table}

\subsection{Comparison to State-of-the-Arts}
After exploring of our proposed model, we compare it with models published on the most recent conferences. For a fair comparison, we use the standard motion and appearance feature extractions used by most models. Reinforcement learning and audio features are not used.
The baseline model is a standard hierarchical-LSTMs. We list the results of the fixed retriever model and jointly trained retriever model. For MSR-VTT, we choose top-$3/10$ retrieved sentences for training/inference. For VATEX, top-$3/10$ sentences are selected for training/inference. Both the pre-training of the retriever and the fine-tuning of RCG are performed on the training set. In Tab.\ref{tab:tab6}, compared with the baseline, RCG achieves remarkable improvements, which proves the effectiveness of our method. Moreover, it outperforms ORG-TRL model even without fine-grained object features and external knowledge, which obtains 3.9\% and 15.7\% relative gains on CIDEr metric for MSR-VTT and VATEX. We have a comparable performance for the other metrics of MSR-VTT and have achieved the best results in VATEX. In addition, compared to MSRVTT, the improvement in VATEX is particularly obvious. It may be because the videos in VATEX are collected by categories and have a large number of videos with the same semantics. This is more consistent with our motivation and practical industrial applications. 

\subsection{Qualitative Analysis}
We visualize the heatmap of the words copied from the retrieved sentences and their probabilities during the generation process, as illustrated in Fig.\ref{fig:fig3}. According to the heatmap, whether the words come from the retrieved sentences and each sentence's contribution can be seen intuitively. In Fig.\ref{fig:fig5}, we also visualize the attention weights of retrieved sentences during the process of video-text retrieval. Most of the weights are focused on keywords, \eg, \textit{``scuba diver"}, \textit{``ocean water"} and \textit{``masks"}, which proves the effectiveness of our retriever. Comparing the baseline model, the proposed RCG can acquire more diverse expressions, \eg, ``\textit{breathing out of their masks}" benefit from the coping mechanism, and correct the expressions from ``\textit{a game of frisbee}" to ``\textit{playing catch}" under the guidance of retrieved sentences.

\section{Conclusion}
In this paper, we have presented the RCG for open-book video captioning. RCG efficiently retrieves video-content-relevant sentences from text corpus through a cross-modal retriever, jointly copies cues from multi-retrieved sentences and generates through a copy-mechanism caption generator, and is optimized in a separately or end-to-end manner. Adequate experiments and superior results on two large-scale video caption datasets demonstrate the advantages of our method. Our results suggest that it is practical to copy knowledge not limited to retrieved sentences, \eg, video subtitles, text from video and text from speech \etc, for comprehensive information acquisition and better generation. In the future, we will further explore the content above and use more advanced retrieval to improve model efficiency.

\textbf{Acknowledgment}
This work is supported by the National Key R\&D Plan (No.2018AAA0102802, 2018AAA0102803,2018AAA0102800,2018YFC0823003, 2017YFB1002801), 
Natural Science Foundation of China (No.62036011, 61721004, 61772225, 61972397), 
National Natural Science Foundation of China (No.U2033210),
Key Research Program of Frontier Sciences, CAS (No.QYZDJ-SSW-JSC040), 
National Natural Science Foundation of Guangdong (No.2018B030311046),
Beijing Natural Science Foundation (No.L182058)

{\small
	\bibliographystyle{ieee_fullname}
	\bibliography{egbib}
}
\end{document}